\documentclass[letterpaper, 10 pt, conference]{ieeeconf}  

\IEEEoverridecommandlockouts                              

\title{\LARGE \bf
Sequence-of-Constraints MPC: Reactive Timing-Optimal Control of Sequential Manipulation
}

\author{Marc Toussaint$^{1,2}$, Jason Harris$^1$, Jung-Su Ha$^1$, Danny Driess$^{1,2}$, Wolfgang H{\"o}nig$^1$%
\thanks{The research has been supported by the German Research Foundation (DFG) under Germany’s Excellence Strategy
EXC 2002/1–390523135 ``Science of Intelligence''}
\thanks{$^1$Learning \& Intelligent Systems Lab, TU Berlin}%
\thanks{$^2$Science of Intelligence Excellence Cluster, TU Berlin}%
\thanks{~{\tt\small toussaint@tu-berlin.de}}%
}

  \usepackage{amsmath}
  \usepackage{amssymb}
  \usepackage{amsfonts}
  \allowdisplaybreaks

  \usepackage{amsthm}
  \usepackage{graphicx}
  \usepackage{subcaption}
  
  \usepackage{algorithm}
\usepackage{algpseudocode}
\algrenewcommand{\algorithmicrequire}{\textbf{Input:}}
\algrenewcommand{\algorithmicensure}{\textbf{Maintained from last cycle:}}
\algrenewcommand{\algorithmiccomment}[1]{~\hfill~\hspace*{-5ex}\textit{// #1}}

  \renewcommand{\a}{\alpha}
  \renewcommand{\b}{\beta}
  \renewcommand{\d}{\delta}
    
    \newcommand{\e}{\epsilon}

  \renewcommand{\l}{\lambda}
  \renewcommand{\L}{\Lambda}
    \newcommand{\m}{\mu}

  \renewcommand{\k}{\kappa}
  \renewcommand{\o}{\omega}
  \renewcommand{\O}{\Omega}

  \renewcommand{\r}{\varrho}
    \newcommand{\s}{\sigma}
  
  \renewcommand{\t}{\theta}

  \renewcommand{\SS}{{\cal S}}

  \newcommand{\RRR}{{\mathbb{R}}}

  \renewcommand{\[}{\Big[}
  \renewcommand{\]}{\Big]}

  \renewcommand{\|}{\,|\,}
  
  \renewcommand{\=}{\!=\!}

  \newcommand{\feed}{\nonumber \\}
  \newcommand{\comma}{~,\quad}

  \newcommand{\st}{~~\text{s.t.}~~}

  \newcommand{\half}{\ensuremath{\frac{1}{2}}}

  \renewcommand{\vec}{\boldsymbol}

  \newcommand{\mat}[3][.9]{
    \renewcommand{\arraystretch}{#1}{\scriptscriptstyle{\left(
      \hspace*{-1ex}\begin{array}{#2}#3\end{array}\hspace*{-1ex}
    \right)}}\renewcommand{\arraystretch}{1.2}
  }

  \providecommand{\href}[2]{{\color{blue}USE PDFLATEX!}}
  \providecommand{\url}[2]{\href{#1}{{\color{blue}#2}}}
  \newcommand{\anchor}[2]{\begin{picture}(0,0)\put(#1){#2}\end{picture}}

  \newcommand{\fullhide}[1]{\message{^^JHIDE--Warning (hidden)!^^J}}
  \newcommand{\mytitle}[1]{\title{#1}\newcommand{\thetitle}{#1}}
  \newcommand{\header}{\begin{document}\mytitle\cleardefs}
  \newcommand{\contents}{{\tableofcontents}\renewcommand{\contents}{}}
  \newcommand{\footer}{\small\bibliography{marc,bibs}\end{document}}
  \newcommand{\widepaper}{\usepackage{geometry}\geometry{a4paper,hdivide={25mm,*,25mm},vdivide={25mm,*,25mm}}}
  \newcommand{\moviex}[2]{\movie[externalviewer]{#1}{#2}} 
  \newcommand{\rbox}[1]{\fboxrule2mm\fcolorbox[rgb]{1,.85,.85}{1,.85,.85}{#1}}
  \newcommand{\mpage}[2]{{\begin{minipage}{#1\columnwidth}#2\end{minipage}}}
  \newcommand{\redbox}[2]{\fboxrule1mm\fcolorbox[rgb]{1,.7,.7}{1,.7,.7}{\begin{minipage}{#1\columnwidth}\center#2\end{minipage}}}
  \newcommand{\onecol}[2]{
    \begin{minipage}[c]{#1\columnwidth}#2\end{minipage}}
  \newcommand{\twocol}[5][0]{
    \begin{minipage}[c]{#2\columnwidth}#4\end{minipage}\hspace*{#1\columnwidth}%
    \begin{minipage}[c]{#3\columnwidth}#5\end{minipage}}
  \newcommand{\threecol}[7][0]{%
    \begin{minipage}[c]{#2\columnwidth}#5\end{minipage}\hspace*{#1\columnwidth}%
    \begin{minipage}[c]{#3\columnwidth}#6\end{minipage}\hspace*{#1\columnwidth}%
    \begin{minipage}[c]{#4\columnwidth}#7\end{minipage}}
  \newcommand{\threecoltext}[7][c]{
    \begin{minipage}[#1]{#2\textwidth}#5\end{minipage}%
    \begin{minipage}[#1]{#3\textwidth}#6\end{minipage}%
    \begin{minipage}[#1]{#4\textwidth}#7\end{minipage}}
  \newcommand{\threecoltop}[7][0]{%
   \begin{minipage}[t]{#2\columnwidth}#5\end{minipage}\hspace*{#1\columnwidth}%
   \begin{minipage}[t]{#3\columnwidth}#6\end{minipage}\hspace*{#1\columnwidth}%
   \begin{minipage}[t]{#4\columnwidth}#7\end{minipage}}
  \newcommand{\fourcol}[9][0]{%
   \begin{minipage}[c]{#2\columnwidth}#6\end{minipage}\hspace*{#1\columnwidth}%
   \begin{minipage}[c]{#3\columnwidth}#7\end{minipage}\hspace*{#1\columnwidth}%
   \begin{minipage}[c]{#4\columnwidth}#8\end{minipage}\hspace*{#1\columnwidth}%
   \begin{minipage}[c]{#5\columnwidth}#9\end{minipage}}
  \newcommand{\helvetica}[4]{\setlength{\unitlength}{1pt}\fontsize{#1}{#1}\linespread{#2}\usefont{OT1}{phv}{#3}{#4}}
  \newcommand{\helve}[1]{\helvetica{#1}{1.5}{m}{n}}
  \newcommand{\german}{\usepackage[german]{babel}\usepackage[utf8]{inputenc}}
\newcommand{\sans}{\renewcommand{\familydefault}{\sfdefault}}

\newcommand{\norm}[1]{|\!|#1|\!|}
\newcommand{\expr}[1]{[\hspace{-.2ex}[#1]\hspace{-.2ex}]}

\newcommand{\Jwi}{J^\sharp_W}
\newcommand{\THi}{T^\sharp_H}
\newcommand{\Jci}{J^\natural_C}
\newcommand{\hJi}{{\bar J}^\sharp}
\renewcommand{\|}{\,|\,}
\renewcommand{\=}{\!=\!}
\newcommand{\myminus}{{\hspace*{-.0pt}\text{\rm -}\hspace*{-.5pt}}}
\newcommand{\myplus}{{\hspace*{-.0pt}\text{\rm +}\hspace*{-.5pt}}}
\newcommand{\1}{{\myminus1}}
\newcommand{\2}{{\myminus2}}
\newcommand{\3}{{\myminus3}}
\newcommand{\mT}{{\text{\rm -}\hspace*{-1pt}\top}}
\newcommand{\po}{{\myplus1}}
\newcommand{\pt}{{\myplus2}}
\newcommand{\T}{{\!\top\!}}
\newcommand{\xT}{{\underline x}}
\newcommand{\uT}{{\underline u}}
\newcommand{\zT}{{\underline z}}
\newcommand{\Sum}{\textstyle\sum}
\newcommand{\Int}{\textstyle\int}
\newcommand{\Prod}{\textstyle\prod}

\newenvironment{centy}{
\vspace{15mm}
\large
\hspace*{5mm}
\begin{minipage}{8cm}\it\color{blue}
}{
\end{minipage}
}

\newcommand{\old}{{\text{old}}}
\newcommand{\new}{{\text{new}}}
\newcommand{\MAP}{{\text{MAP}}}
\newcommand{\ML}{{\text{ML}}}

\newcommand{\redArrow}{\quad\anchor{0,-1}{\includegraphics[scale=.5]{figs/redArrow}}}
\newcommand{\pub}[1]{{\color{green}\helvetica{8}{1.3}{m}{n}#1\\}}
\DeclareMathOperator{\opKL}{KL}
\newcommand{\KL}[2]{\opKL\big(#1\,\big|\!\big|\,#2\big)} 

\renewcommand{\show}[2][.7]{\centerline{\includegraphics[width=#1\columnwidth]{#2}}}
\newcommand{\showh}[2][.8]{\includegraphics[width=#1\columnwidth]{#2}}
\newcommand{\shows}[2][.8]{\centerline{\includegraphics[scale=#1]{#2}}}
\newcommand{\showhs}[2][.8]{\includegraphics[scale=#1]{#2}}
\newcommand{\mov}[2]{\movie[externalviewer]{{\color{blue}\small #1}}{movies/#2}}
\newcommand{\movex}[2]{\movie[externalviewer]{#1}{#2}} 
\newcommand{\movh}[3][autostart,loop]{
\movie[#1]{\showh[#2]{#3.jpg}}{#3.mp4}%
\movie[externalviewer]{$\circ$}{#3.mp4}
}
\newcommand{\movc}[3][autostart,loop]{\centerline{\movh[#1]{#2}{#3}}}
\newcommand{\mymovie}[3][]{\centerline{\movie[autostart,loop,#1]{\includegraphics[#1]{#2}}{#3}}}

\newcommand{\cen}[1]{\centerline{#1}}

\newcommand{\citing}[1]{
{\color{citcol}\ttiny#1\par}
}

\newcommand{\cit}[3]{
\par\smallskip
{\color{citcol}\ttiny #1: \emph{#2}. #3 \par}
}

\newcommand{\citurl}[4]{
\par\smallskip
{\color{citcol}\ttiny #1: \protect{\href{#4}{\color{blue}{#2.}}} #3 \par}
}

\newcommand{\cito}[3]{
\par\smallskip
{\color{bluecol}\ttiny #1: \emph{#2}. #3 \par}
}

\newcommand{\redoMacrosInProof}{
  \renewcommand{\d}{\delta}
  \renewcommand{\=}{\!=\!}
}

\newcommand{\pdflatex}{
  \definecolor{bluecol}{rgb}{0,0,.5}
  \usepackage[
    colorlinks,
    urlcolor=bluecol,
    citecolor=black,
    linkcolor=bluecol,
    pdfborder={0 0 0},
    pdfpagemode=UseOutlines, 
    pdfauthor={Marc Toussaint}
  ]{hyperref}
  \DeclareGraphicsExtensions{.pdf,.png,.jpg,.eps}
  \renewcommand{\r}{\varrho}
  \renewcommand{\l}{\lambda}
  \renewcommand{\L}{\Lambda}
  \renewcommand{\s}{\sigma}
  \renewcommand{\b}{\beta}
  \renewcommand{\d}{\delta}
  \renewcommand{\k}{\kappa}
  \renewcommand{\t}{\theta}
  \renewcommand{\O}{\Omega}
  \renewcommand{\o}{\omega}
  \renewcommand{\SS}{{\cal S}}
  \renewcommand{\=}{\!=\!}
}

\newenvironment{code}{%
\codefont
\begin{shaded}
}{
\end{shaded}
}

\newcommand{\signature}{
  \includegraphics[height=6ex]{signatureColor} 
}

\usepackage{etoolbox}

\usepackage{multirow}
\usepackage{colortbl}
\usepackage{empheq}

\newcommand{\mypause}{\pause}
\newcommand{\defi}[1]{\textbf{#1}}
\newcommand{\red}[1]{\emph{\color{red}#1}}
\newcommand{\pos}{{\textsf{pos}}}
\newcommand{\eff}{{\textsf{eff}}}
\newcommand{\rot}{{\textsf{rot}}}
\newcommand{\veC}{{\textsf{vec}}}
\newcommand{\quat}{{\textsf{quat}}}
\newcommand{\col}{{\textsf{col}}}
\newcommand{\de}[2]{\frac{\partial #1}{\partial #2}}
\newcommand{\target}{{\text{target}}}
\newcommand{\near}{{\text{near}}}
\newcommand{\qfree}{Q_{\text{free}}}
\renewcommand{\vec}{\boldsymbol}
\newcommand{\lft}{\text{left}}
\newcommand{\rgh}{\text{right}}
\DeclareMathOperator{\real}{real}
\newcommand{\prev}{{\text{prev}}}
\newcommand{\TR}[2]{T_{{#1}\to{#2}}}
\newcommand{\RO}[2]{R_{{#1}\to{#2}}}
\newcommand{\liter}{\helvetica{8}{1.1}{m}{n}\parskip 1ex}
\newcommand{\Fc}{\color{green}F}
\newcommand{\muc}{\color{blue}\mu}
\newcommand{\Astar}{A$^*$}

\providecommand{\defn}[1]{\textbf{#1}}

\newcommand{\adec}{\r_\a^-}
\newcommand{\ainc}{\r_\a^+}
\newcommand{\ldec}{\r_\l^-}
\newcommand{\linc}{\r_\l^+}
\newcommand{\minc}{\r_\m^+}
\newcommand{\mdec}{\r_\m^-}
\newcommand{\step}{{\delta}}
\newcommand{\lsstop}{\r_{\text{ls}}}
\newcommand{\stepmax}{\step_{\text{max}}}
\newcommand{\Min}{\text{min}}
\newcommand{\Max}{\text{max}}

\definecolor{boxcol}{rgb}{.85,.9,.92}
\newcommand{\eqbox}[1]{\centerline{\fboxrule0mm\fcolorbox{boxcol}{boxcol}{#1}}}
\newcommand{\movgb}[1]{\hfill\movie[externalviewer]{\small[movie]}{/home/mtoussai/movies/10-goalDirectedBehavior/#1}}
\newcommand{\demo}[1]{{{\color{blue}[\small #1]}}}

\graphicspath{{../pics-robotics/}{../pics-ML/}{../pics-all/}{../pics-all2/}{../pics-Optim/}}
\DeclareGraphicsExtensions{.pdf,.png,.jpg}

\newcommand{\SUM}{\texttt{sum}}
\usepackage{float}

\makeatletter
\newcommand{\NewParNoBreak}[1][\parskip]{\par\vspace*{-\parskip}\vspace*{#1}\nobreak\@afterheading}
\makeatother

\newtheorem{property}{Property}

\newcommand{\goal}{{\text{goal}}}
\newcommand{\switch}{{\text{switch}}}
\newcommand{\Osos}{\texttt{S}}
\newcommand{\Oineq}{\texttt{I}}
\newcommand{\Oeq}{\texttt{E}}
\newcommand{\Oc}{\texttt{C}}
\newcommand{\ff}{{\mathcal{f}}}
\newcommand{\secmpc}{{\sc SecMPC}}
\newcommand{\rt}{{\mathcal{T}}}

\newcommand{\marc}[1]{{\textbf{marc: #1}}}

  \definecolor{bluecol}{rgb}{0,0,.5}
  \usepackage[
    colorlinks,
    urlcolor=bluecol,
    citecolor=black,
    linkcolor=bluecol,
    pdfborder={0 0 0},
    pdfpagemode=UseOutlines, 
    pdfauthor={Marc Toussaint}
  ]{hyperref}
  \DeclareGraphicsExtensions{.pdf,.png,.jpg,.eps}
  \renewcommand{\r}{\varrho}
  \renewcommand{\l}{\lambda}
  \renewcommand{\L}{\Lambda}
  \renewcommand{\s}{\sigma}
  \renewcommand{\b}{\beta}
  \renewcommand{\d}{\delta}
  \renewcommand{\k}{\kappa}
  \renewcommand{\t}{\theta}
  \renewcommand{\O}{\Omega}
  \renewcommand{\o}{\omega}
  \renewcommand{\SS}{{\cal S}}
  \renewcommand{\=}{\!=\!}

\begin{document}

\maketitle
\thispagestyle{empty}
\pagestyle{empty}

\maketitle

\begin{abstract}
Task and Motion Planning has made great progress in solving hard
sequential manipulation problems. However, a gap between such planning
formulations and control methods for reactive execution remains. In
this paper we propose a model predictive control approach dedicated to
robustly execute a single sequence of constraints, which corresponds
to a discrete decision sequence of a TAMP plan. We decompose the
overall control problem into three sub-problems (solving for
sequential waypoints, their timing, and a short receding horizon path)
that each is a non-linear program solved online in each MPC cycle. The
resulting control strategy can account for long-term interdependencies
of constraints and reactively plan for a timing-optimal transition
through all constraints. We additionally propose phase backtracking
when running constraints of the current phase cannot be fulfilled,
leading to a fluent re-initiation behavior that is robust to
perturbations and interferences by an experimenter.
\end{abstract}


\section{Introduction}

Task and Motion Planning (TAMP)
\cite{2021-garrett-IntegratedTaskMotion} has made great progress in
recent years in solving hard sequential manipulation
problems. However, to bring such plans to reactive execution remains a
challenge and typically relies on a predefined set of controller
primitives to execute individual actions
\cite{2019-paxton-RepresentingRobotTask}. As an example for robust
execution, Boston Dynamics has been showcasing several demonstrations
where humans massively perturb a linear plan, not only in locomotion
but also sequential
manipulation.\footnote{\url{https://youtu.be/aFuA50H9uek}} Candidates
to design such highly reactive and robust behavior include hierarchies
of convergent controllers (funnels)
\cite{2017-majumdar-FunnelLibrariesRealtime}, and the use of state
machines to orchestrate transitioning between control modes
\cite{2000-egerstedt-BehaviorBasedRobotics,2016-eppner-LessonsAmazonPicking}. However,
the latter again relies on a predefined set of controllers per
action. We aim for a general approach to more directly translate a
TAMP plan to a reactive execution strategy.


\begin{figure}
  \show[.7]{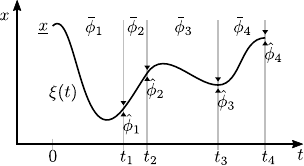}
  \caption{\label{figSec}Illustration of the Sequence-of-Constraints
    Optimal Control problem (\ref{eqSecCtrl}): We optimize a path
    $\xi(t)$ and timings $t_{1:K}$ subject to waypoint constraints
    $\hat\phi_{1:K}$ and running constraints
    $\bar\phi_{1:K}$. Constraints concern any non-linear
    differentiable features $\phi$ of the system configuration and/or
    velocity. The system configuration $\xi(t)$ includes manipulation
    dofs such as relative grasp, push and placement poses, which imply
    interdependencies between waypoint constraints (see Section
    \ref{secDepend}).  }
\end{figure}

In the context of trajectory optimization, the translation from
trajectory planning to control is immediate via repeated online
planning, in particular model-predictive control (MPC) approaches. Here, the
same underlying problem formulation (in terms of the path objectives)
equally allows to derive optimal plans as well as robust control
strategies. Given existing optimization-based formulations of TAMP
\cite{2015-toussaint-LogicgeometricProgrammingOptimizationbased,2020-stouraitis-MultimodeTrajectoryOptimization,2020-stouraitis-OnlineHybridMotion},
it might seem that reactive execution could easily be derived from the
same underlying optimality formulation by a standard MPC approach
\cite{2020-ha-ProbabilisticFrameworkConstrained}. However, two core
challenges arise: 1) To achieve correct sequential manipulation, we
require instant reactiveness to perturbations that concern constraints
in the far future of the plan. For instance, when the target at the
end of a pushing maneuver is perturbed, this might require an instant
reaction in the approach angle in an early phase of the
maneuver. Correctly accounting for such long-term interdependencies and
coupling them to immediate reactiveness is beyond standard
(e.g.\ fixed-horizon) MPC approaches and existing TAMP execution
methods
\cite{2019-paxton-RepresentingRobotTask,2020-stouraitis-OnlineHybridMotion,2020-migimatsu-ObjectcentricTaskMotionb}. 2)
TAMP plans include switches in kinematic and dynamic constraints
\cite{2018-toussaint-DifferentiablePhysicsStable}, and the original
plan might provide a temporal scheduling of such switches. However,
in a reactive execution approach also the timing of these constraint switches needs
to be reactive. This either requires employing optimal control methods
invariant to such switches
\cite{2014-posa-DirectMethodTrajectory,2012-mordatch-DiscoveryComplexBehaviors}
or to include explicit timing-estimation and -optimization as part of
the MPC problem.

This paper is dedicated to address these two challenges. We assume
that a TAMP plan was computed and is given in terms of a linear
sequence of constraints (aka.\ ``skeleton''). Plan switching or
re-planning action sequences \cite{2019-paxton-RepresentingRobotTask}
is beyond the scope of this work. Instead we focus on deriving a
Timing-Optimal Sequence-of-Constraints MPC (\secmpc) approach to
robustly control through a sequence of constraints that are subject to
perturbations during execution, which includes a backtracking mechanism
when perturbations break the running constraints of a manipulation
phase. Backtracking here means that we stay within the linear
sequence of constraints, but back up to an earlier phase whenever
constraints of the current phase are not fulfilled. This leads to an
automatic reactive re-initiation of parts of the sequence.

Our MPC approach includes the timing of constraints as a decision
variable, with the objective being a combination of total time and
control costs, which we call \emph{timing-optimal}.\footnote{Note that
  the term \emph{time-optimal} is typically used when the objective is
  \emph{only} total time (typically with limits on controls), leading
  to bang-bang type solutions; we specifically use the term
  \emph{timing-optimal} to emphasize that we optimize timing for a
  combination of total time and control costs.} However, unlike
standard MPC approaches our formulation is not constrained to a short
receding horizon, but instead approximates the full horizon problem to
ensure timing and waypoint consistency, but with varying
resolution. More specifically, each MPC cycle includes solving for 1)
future waypoints to be consistent with future constraints and their
dependencies in the sequence, 2) the timing of waypoints, and 3) a
short receding horizon path to fulfill immediate collision and path
constraints.
%
The resulting overall system defines a \emph{temporally consistent}
controller, where, if there are no perturbations, the time-to-go and
future path estimates of two consecutive MPC cycles are consistent
modulo the shift in time, guaranteeing the conclusion of the full
sequence in the originally estimated timing and along the originally
estimated path.



We demonstrate the control framework on a pick-and-place and quadrotor
scenario, but in particular on a pushing-with-stick scenario. The latter
experiment is inspired by the observation in
\cite{2018-toussaint-DifferentiablePhysicsStable} that humans execute
pushing tasks drastically different from optimal stable pushing plans:
The overall behavior is dominated by frequent re-initiation of brief
pushing phases, where each push is instable and rather imprecise, but
the re-initiations robustly control the object to the
target.\footnote{See the data at the end of
  \url{https://youtu.be/-L4tCIGXKBE}.} Our approach can be seen as a
response to this observation, providing a model of how a
sequence-of-constraints plan can lead to complex behavior reactively
cycling through execution phases.



\section{Related work}

\subsection{Bridging between TAMP and Reactive Execution}

In \cite{2019-paxton-RepresentingRobotTask}, a reactive sequential
manipulation framework is proposed that can reactively replan logic
decisions, but requires a manual design of the controllers per logic
action. \cite{2021-li-ReactiveTaskMotion} similarly combine linear
temporal logic TAMP planning with behavior trees for reactive action
selection and plan switching under interventions. Both works go beyond
our work in terms of the reactiveness on the action level, but do not
consider a coherent optimal control formulation, timing-optimality, or
the interdependencies of future waypoints in the fluent execution of a
fixed skeleton.

To enable reactive execution of TAMP plans,
\cite{2020-migimatsu-ObjectcentricTaskMotionb} proposes to interpret
them in object-centric Cartesian coordinates and design corresponding
perception-based operational space controllers for
execution. \cite{2019-schmitt-ModelingPlanningManipulation} provides a
novel method to automatically design a sequence of controllers
using kino-dynamic planning. In contrast, our approach is a pure
optimization-based MPC approach that can take the interdependencies of
future waypoints into account.

Receding-Horizon TAMP
\cite{2021-castaman-RecedingHorizonTask,2021-braun-RHHLGPRecedingHorizon,2020-hartmann-RobustTaskMotion}
has been proposed to speed up and decompose long-horizon TAMP
problems, but without bridging to low-level reactive control.

\subsection{Timing Optimization \& MPC}

The timing of future constraints is a central concern in our approach,
that has previously been studied in the context of locomotion
\cite{2017-caron-WhenMakeStep}. In the context of quadrotor flight,
\cite{2018-lin-EfficientTimeoptimalTrajectory} proposed an efficient
time-optimal trajectory generation method, and
\cite{2021-foehn-TimeoptimalPlanningQuadrotor} proposed an interesting
alternative formulation based on progress variables which ensure that
an optimal path transitions through a given sequence of constraints in
a timing-optimal manner.
\cite{2017-pham-StructureTimeoptimalPath,2018-pham-NewApproachTimeoptimal}
proposed further novel methods for time-optimal path
planning. However, these methods are used for trajectory
optimization rather than as a reactive control framework.


In the context of MPC, \cite{2022-bhardwaj-STORMIntegratedFrameworka}
proposed a sampling-based MPC method for robotic manipulation, and
\cite{2017-rosmann-TimeoptimalNonlinearModel,2011-vandenbroeck-ModelPredictiveControl}
further advanced the state in time-optimal MPC, but these works do not
consider controlling through a sequence of switching constraints as
they appear in TAMP plans.

Finally, hybrid control, and in particular trajectory optimization
through switching constraints \cite{2014-posa-DirectMethodTrajectory,2022-moura-NonprehensilePlanarManipulation}
have in principle high potential for MPC through manipulation, but
have to our knowledge not been extended to reactively account for
interdependencies of future waypoint constraints and their timing.



\section{Problem Formulation}

\subsection{Sequence-of-Constraints Optimal Control}

We consider a system configuration space $X=\RRR^n$ which includes
robot dofs as well as manipulation dofs such as relative grasp, push,
and place poses. The latter typically underlies constancy constraints
that imply long term dependencies \cite{2018-toussaint-DifferentiablePhysicsStable}, and we treat them in analogy to constant design parameters \cite{2021-toussaint-CoOptimizingRobotEnvironment}. We use $\underline x = (x,\dot x)$ to denote the system
state, with configuration $x\in X$ and velocity $\dot x \in \RRR^n$. A
Sequence-of-Constraints optimal control problem is specified by a
tuple
$$(\hat\phi_{1:K}, \bar\phi_{1:K}) ~,$$ which is a sequence of $K$
waypoint constraint functions $\hat\phi_k: X \to \RRR^{\hat d_k}$, as
well as $K$ running constraint functions $\bar\phi_k: X\times\RRR^n
\to \RRR^{\bar d_k}$. Each constraint function $\phi$ may have
a different output dimension (number of constraints) and is assumed to
be smoothly differentiable. Fig.~\ref{figSec} illustrates the
notation. The objective is to find a trajectory $\xi:[0,t_K]\to X$
and its timing $t_{1:K}, t_k\in \RRR_+$ to minimize
\begin{subequations}\label{eqSecCtrl}
\begin{align}
  \min_{\xi,t_{1:K}} ~& t_K + \a \int_0^{t_K} c(\xi(t),\dot\xi(t),\ddot \xi(t))~ dt \\
  \st & \xi(0)=x,~ \dot \xi(0)=\dot x,~ \dot\xi(t_K)=0 ~,\\
  &\forall_k: 0 < t_k < t_{k\po} ~,\\
  &\forall_k: \hat\phi_k(\xi(t_k)) \le 0,~ \forall_{t\in[t_{k\1},t_k]}: \bar\phi_k(\underline \xi(t)) \le 0 ~.
\end{align}
\end{subequations}
In this notation, constraint functions $\phi$ define inequalities on
system configurations $x$ and velocities $\dot x$, but this is meant to
include equality constraints.\footnote{In practice, we label outputs
  of $\phi$ to be either equalities or inequalities, so that a solver
  can treat them accordingly, e.g.\ using an Augmented Lagrangian-term
  or a log barrier-term, respectively.} For example, running
constraints can impose dynamics and collision avoidance constraints,
while waypoint constraints can impose constraints for transitioning
between dynamics modes, e.g., constraints to initiate a stable grasp
or push in our experiments. Relating to TAMP formalisms, waypoint
constraints correspond to constraints for switching between modes, while
running constraints encode the dynamics of a mode
\cite{2018-toussaint-DifferentiablePhysicsStable,2019-schmitt-ModelingPlanningManipulation},
which are imposed by a skeleton (logical decision sequence). Finally,
$c(\xi(t),\dot\xi(t),\ddot \xi(t))$ represents control costs scaled with $\a\in\RRR$: In our
experiments we will simply penalize square accelerations $\norm{\ddot
  q}^2$ in robot dofs $q\subseteq x$, but also add a small pose regularization
$\norm{q-q_\text{home}}^2$ that prefers robot poses close to the
homing position and prevents the system from drifting
through null-spaces in long manipulation sequences.

\subsection{Interdependency of waypoint constraints}\label{secDepend}

Sequential robotic manipulation implies long-term dependencies
e.g.\ between the pick and place pose of an object via the constraint
of a constant relative grasp pose, or the onset of a push and the
final placement of the object via stable push constraints.
To account for such dependencies we augment
the system configuration space $X$ to include manipulation dofs. For
instance, such manipulation dofs represent a constant relative grasp
pose or final push-placement. By imposing constancy constraints
(either via running constraints $\bar\phi$ or by explicit sharing of
dofs across time slices, see
\cite{2021-toussaint-CoOptimizingRobotEnvironment}), the sequence of
waypoint constraints $(\hat\phi_1(\xi(t_1)),..,\hat\phi_K(\xi(t_K)))$
concerns parameters shared across waypoints to model their
interdependencies.

\section{Decomposition}

The full problem (\ref{eqSecCtrl}) would raise high computational
costs within an MPC loop: The running constraints concern the
continuous-time path and may require a fine resolution time
discretization, e.g.\ to evaluate collisions. This would imply that we
would have to jointly optimize over a fine resolution path through all
waypoints and their timing, which is tractable offline but costly
within an MPC loop.

We therefore propose to approximate and decompose the full path
problem as follows. First note that a coordinate descent approach
\cite{2015-wright-CoordinateDescentAlgorithms} to (\ref{eqSecCtrl})
would alternate between solving for the timing $t_{1:K}$ given the
current path $\xi$, and vice versa. However, optimizing the
full-horizon fine path $\xi$ within each MPC cycle would still be
computationally inefficient. We therefore decompose the problem
further by representing the path coarsely in terms of the waypoints
$x_{1:k} = \xi(t_1),..,\xi(t_k)$, and finely only in a short receding
horizon of length $H$ ($\approx$1sec). More specifically, this defines
our three main decision variables,
\begin{enumerate}
  \item the timing $t_{1:K}$,  
  \item the waypoints $x_{1:K} = \xi(t_1),..,\xi(t_K)$,
  \item the short receding horizon path $\xi:[0,H] \to X$, in fine time discretization.
\end{enumerate}
For each variable, we define a sub-problem assuming the other
variables fixed, as detailed in the following sections. Each cycle
iterates only once over the three sub-problems, but when in subsequent
MPC cycles the optima become stable we have a joint and consistent
solution. However, this joint solution still is only approximate to
the full problem (\ref{eqSecCtrl}) as the waypoints only coarsely
represent the full path $\xi$.

\subsection{The waypoints problem}

The waypoints problem assumes a fixed timing $t_{1:K}$ and solves for
waypoints $x_{1:K}$ at these prescribed timings. This problem setting
coincides with standard waypoint optimization, in particular the methods
employed previously in the context of optimization-based TAMP: Namely,
in
\cite{2017-toussaint-MultiboundTreeSearch,2018-toussaint-DifferentiablePhysicsStable}
we described the \emph{sequence bound} as optimizing only over
waypoints, but subject to manipulation constraints, e.g.\ that a
relative grasp pose remains stable.  As we discretized the path
representation, the running constraints $\bar\phi$ in
(\ref{eqSecCtrl}) now become constraints that couple two consecutive
waypoints, $\bar\phi_k(x_{k\1},x_k)$, e.g.\ ensuring that the relative grasp in
two consecutive waypoints remains equal. The waypoints
problem therefore is of the form
\begin{align}\label{eqWaypointP}
  \min_{x_{1:K}} ~& \sum_{k=1}^K \tilde c(x_{k\1},x_k) \\
  \st & \forall_k: \hat\phi_k(x_k) \le 0,~ \bar\phi_k(x_{k\1},x_k) \le 0 ~,
\end{align}
where $\tilde c$ subsumes control costs between waypoints, and
which we tackle using a solver from previous work
\cite{2018-toussaint-DifferentiablePhysicsStable}.


\subsection{The timing problem under a cubic spline model}

The timing problem assumes fixed waypoints $x_{1:K}$. To optimize the timing of waypoints we choose the control costs to be squared accelerations, neglecting the above mentioned pose regularization. Therefore, the timing sub-problem aims to find a path $\xi:[0,t_K]\to X$ through the given waypoints and its timing $t_{1:K}, t_k\in \RRR_+$ to minimize
\begin{align}\label{eqTimingOrg}
  \min_{t_{1:K},\xi} ~& t_K + \a \int_0^{t_K} \ddot \xi(t)^2~ dt \\
  \st & \xi(0)=x,~ \dot \xi(0)=\dot x,~ \dot\xi(t_K)=0,~ \\
  & \forall_k: \xi(t_k) = x_k,~ 0 < t_k < t_{k\po}
\end{align}
In consistency with (\ref{eqSecCtrl}) we  minimize for total time $t_K$, but simplify the control costs to smoothness $\ddot \xi^2$ and replace non-linear constraints with waypoint constraints that were assumed to solve for the non-linear constraints.\footnote{Our framework would also allow to impose hard control limits on $\ddot \xi$ and have a pure time-optimal formulation, not mixing with control costs. We found this to be  practical with a log-barrier methods as a solver; however robustly warm-starting log-barrier methods within an MPC cycle turned out a substantial challenge, which is why we decided to back away from hard control limits and a pure time-optimal formulation in this first work on \secmpc.}

For square acceleration costs and given boundary conditions $\xi(t_k), \dot\xi(t_k)$ at each time step, the optimal path (minimizing $\int \ddot \xi$) between two consecutive steps is a cubic polynomial. The optimal solution to the above problem is therefore a piece-wise cubic spline, that is parameterized by $t_{1:K}$ and $\xi(t_k), \dot\xi(t_k)$, where $\xi(t_k)=x_k$ is fully known. Defining $v_k= \dot\xi(t_k)$ we therefore can rewrite the timing problem using the decision variables $t_{1:K}, v_{1:K\1}$, or alternatively, $\tau_{1:K}, v_{1:K\1}$ with the delta timings $\tau_{1:K}$ such that
\begin{align}
  t_k = \textstyle\sum_{i=1}^k \tau_k ~.
\end{align}

We define the cubic piece cost as
\begin{align}
  \psi(x_0,v_0,x_1,v_1,\tau) &~= \min_{z} \int_0^\tau \ddot z(t)^2~ dt \feed
  \st& \mat{c}{z(0)\\\dot z(0)}=\mat{c}{x_0\\v_0},\mat{c}{z(\tau)\\\dot z(\tau)}=\mat{c}{x_1\\v_1}~. \label{eqLeap}
\end{align}
For boundary conditions $(x_0, v_0, x_1, v_1, \tau)$, this is solved by a cubic spline $z(t) = a t^3 + b t^2 + c t + d$ with
\begin{align}
d &= x_0 \comma c = \dot x_0 \\
b &= \frac{1}{\tau^2}\[  3(x_1-x_0) - \tau(\dot x_1 + 2 \dot x_0) \]\\
a &= \frac{1}{\tau^3}\[ -2(x_1-x_0) + \tau(\dot x_1 + \dot x_0) \] ~,
\end{align}
and the minimal cost $\psi$ of (\ref{eqLeap}) is
\begin{align}
\psi
&= \int_0^\tau \ddot z(t)^2~ dt
 = 4 \tau b^2  + 12 \tau^2 ab + 12 \tau^3 a^2 \\
&= \frac{12}{\tau^3}~[(x_1 - x_0)-\frac{\tau}{2}(v_0+v_1)]^2+\frac{1}{\tau}(v_1-v_0)^2 \\
&= \frac{12}{\tau^3} D^\T D + \frac{1}{\tau} V^\T V = \tilde D^\T \tilde D + \tilde V^\T \tilde V \label{eqLeapSOS}\\
& D := (x_1 - x_0)-\frac{\tau}{2}(v_0+v_1),~ V:=v_1-v_0,~ \\
& \tilde D := \sqrt{12}~ \tau^{-\frac{3}{2}}~ D,~ \tilde V := \tau^{-\half}~ V ~.
\end{align}
Here, (\ref{eqLeapSOS}) expresses the cubic piece cost as a
least-squares of differentiable features of
$(x_0,v_0,x_1,v_1,\tau)$, where $D$ can be interpreted as distance
to be covered by accelerations, and $V$ as necessary total
acceleration. The Jacobians of $\tilde D$ and $\tilde V$ w.r.t.\ all
boundary conditions are trivial. Exploiting the least-squares
formulation of $\psi$ we can use the Gauss-Newton approximate Hessian.

Therefore, the timing sub-problem (\ref{eqTimingOrg}) is equivalent to optimizing over the boundary conditions of cubic pieces,
\begin{align}\label{eqTimingP}
  \min_{\tau_{1:K},v_{1:K\1}} \sum_{k=1}^K \tau_K + \a \sum_{k=1}^K \psi(x_{k\1}, v_{k\1}, x_k, v_k, \tau_k) ~,
\end{align}
where $x_0=x$, $v_0=\dot x$ and $v_K = 0$. Note that in (\ref{eqTimingOrg}) and also the
original problem (\ref{eqSecCtrl}) we required $\dot\xi(t_K)=0$ at the
end of the manipulation, which is why we do not have a decision
variable $v_K$ for the last waypoint.

Due to the least-squares nature of all objectives, the timing
sub-problem (\ref{eqTimingP}) is efficient to solve for and
yields a timing as well as waypoint velocities $v_k$, which
together define a piece-wise cubic spline $\xi^*$ through the optimized
waypoints. While the waypoints problem and the short horizon problem
below can be tackled by standard trajectory optimization methods used
in previous work, we transcribed the timing problem into a non-linear
least squares formulation and employed a sparse Gauss-Newton method to
solve it.







\subsection{The short receding horizon problem}

Finally, as is standard in MPC approaches, we solve for a short
receding horizon path that is more finely discretized, e.g.\ to
account for collisions. Let $\xi^*$ be the cubic spline that results
from the timing solution based on the previous waypoint
solutions. This $\xi^*$ incorporates the full-horizon information from
the constraints. We formulate the short horizon MPC problem to track
this reference modulo fulfilling running constraints. Therefore, if no
running constraints (e.g.\ collisions) are active, the short horizon
MPC will reproduce the reference $\xi^*$.

Specifically, we solve for a path $\xi^H: [0,H] \to X$ to minimize
\begin{align}\label{eqShortPathP}
  \min_\xi ~& \int_0^H \a~ \ddot\xi(t)^2 + \norm{\xi(t) - \xi^*(t)}^2~ dt \\
  \st & \xi(0)=x,~ \dot \xi(0)=\dot x ~,\\
  &\forall_{t\in[0,H]}: \bar\phi_{k(t)}(\underline \xi(t)) \le 0 ~,
\end{align}
which is the short horizon version of the original problem
(\ref{eqSecCtrl}), but replacing waypoint constraints and timing by
the reference tracking cost $\norm{\xi^H(t) - \xi^*(t)}^2$.

We use standard trajectory optimization to solve the short horizon problem,  time-discretizing the horizon $[0,H]$. Note that this time interval may ``cross'' a waypoint and corresponding waypoint constraint. However, as the timing $t_{1:K}$ is given and the reference $\xi^*(t)$ is adapted to cross the waypoint with an optimal timing, the short horizon solution will aim to adopt this timing and we have a clear time embedding of the time-discretization.

\section{Sequence-of-Constraints MPC}


Algorithm \ref{secmpc} describes the concrete MPC cycle we propose and
evaluate in our experiments. The core of each cycle is to iteratively
solve the three sub-problems described in the previous section, which
are indicated with bold comments in the pseudo code. However, as the
pseudo code indicates, there are additional details that concern phase
management and stability that we found essential to have a practical
MPC system, and are discussed in the following. We use the term
\emph{phase} to denote the integer $\k\in\{1,..,K\}$ that indicates
which future constraints $\phi_{\k:K}$ are remaining.



\newcommand{\xv}{{\underline x}}
\begin{algorithm}[t]
  \caption{\label{secmpc}
    \secmpc{} Cycle}
  \begin{algorithmic}[1]
    \Ensure last time $\rt'$, phase $\k$, waypoints $x_{\k:K}$, delta timings $\tau_{\k:K}$, velocities $v_{\k:K}$, short path $\xi^H$
    \Require state $\xv = (x,\dot x)$ (including external objects) measured at real time $\rt$
    \State $\d \gets \rt - \rt'$ \Comment{real time since last cycle}
    \State $\tau_\k \gets \tau_\k -\d$ \Comment{shift timing}
    \If{$\tau_\k<0$} \Comment{expected transition}
    \If{$\norm{\hat\phi_\k(\xv)}\le\hat\t$}
    \State \textbf{If} $\k<K$ \textbf{then} $\k \gets \k+1$ \Comment{phase progression} \label{codeWaypointFails}
    \Else
    \State $\tau_\k \gets \tau_\text{init}$
    \EndIf
    \EndIf
    \While{$\norm{\bar\phi_\k(\xv)}>\bar\t$ \textbf{and} $\k>1$} \Comment{phase backtracking}  \label{codeRunningFails}
    \State $\tau_{\k\1},\tau_\k \gets \tau_\text{init}$, $\k\gets \k-1$
    \EndWhile
    \State $\xv \gets$ Filter($\xv, \xi^*(\rt), \dot\xi^*(\rt)$) \label{codeFilter}
    \State $x_{\k:K} \gets$ solve(\ref{eqWaypointP}) \Comment{\textbf{waypoints}, given $\xv, \tau_{\k:K}$}
    \If{$\tau_\k>\e$} \label{codeCutoff}
    \State $\tau_{\k:K},v_{\k:K} \gets$ solve(\ref{eqTimingP}) \Comment{\textbf{timing}, given $x_{\k:K}$}
    \EndIf
    \State $\xi^*\gets$ CubicSpline($\tau_{\k:K},x_{\k:K},v_{\k:K}$)
    \State $\xi^H \gets$ solve(\ref{eqShortPathP}) \Comment{\textbf{short path}, given $\xv, \xi^*$}
    \State $\rt' \gets \rt$, $\rt\gets$ realTime()
    \State Return $\xi^H$ for execution 
  \end{algorithmic}
\end{algorithm}

\subsection{Temporal Consistency of Timing MPC}

We first establish a basic property of our MPC approach, namely
temporal consistency under the assumption of no perturbations, i.e.,
for a deterministic control system and no external object
interventions. Under this assumption, waypoint optimization will
consistently converge to the same waypoints, and for this analysis we
assume the waypoints $x_{1:K}$ are fixed. Further, without perturbations the
deterministic control of $\xi^H$ will coincide with the timing-optimized
cubic spline $\xi^*$, and we have:
\begin{property}[Temporal (Bellman) Consistency]
Consider the solutions $(\tau^{(1)}_{1:K},v^{(1)}_{1:K\1})$ and
$(\tau^{(2)}_{1:K},v^{(2)}_{1:K\1})$ to the MPC problem in two
consecutive MPC iterations with real time gap $\d$. If the system is
deterministic and the expected time to the first waypoint is
$\tau^{(1)}_1 > \d$, it holds
\begin{align}
\tau^{(2)}_1 &= \tau^{(1)}_1 - \d ~, \\
\tau^{(2)}_{2:K} &= \tau^{(1)}_{2:K} ~, \\
v^{(2)}_{1:K\1} &=v^{(1)}_{1:K\1} ~.
\end{align}
That is, the solutions are identical except for the timing
$\tau^{(2)}_1$ of the first waypoint reflecting the real time $\d$
that has passed.
\end{property}
\begin{proof}
  The property is a direct consequence of Bellman's optimality
  principle: Both $(\tau^{(1)}_{1:K},v^{(1)}_{1:K\1})$ and
  $(\tau^{(2)}_{1:K},v^{(2)}_{1:K\1})$ encode the cubic splines
  $\xi^{(1)}$ and $\xi^{(2)}$. Since after the first cycle we execute
  $\xi^{(1)}$ deterministically, Bellman's principle applied to the
  path optimality (\ref{eqTimingOrg}) requires the remaining optimal
  path $\xi^{(2)}$ to coincide with the previously optimal path
  $\xi^{(1)}$.
\end{proof}
%
Further we have the property that, if $\tau^{(1)}_1 \le \d$, the
deterministic system will transition through the first waypoint $x_1$
exactly at time $\tau^{(1)}_1$. When transitioning through a waypoint, the next MPC
solution will be temporally consistent to the previous, in the sense
that $\tau_{2:K}$ becomes the new decision variable and $\tau^{(2)}_2
= \tau^{(1)}_2 - (\d-\tau^{(1)}_1)$ will be reduced by the real time
after waypoint transitioning.

\subsection{Waypoint Transitioning under Stochasticity}\label{secStoch}

In contrast to the above, exact
waypoint transitioning is in principle impossible (of measure
zero) when we have the slightest system stochasticity. Fig.~\ref{figTO}(a) displays timing-optimal paths when the
waypoint is offset for varying start conditions. We see a clear
discontinuity in behavior between directly steering to the waypoint
and deciding to pass the waypoint and looping back. In the stochastic
case, the system will eventually always experience a sufficient offset
and MPC will be stuck in indefinitely looping to re-target the same
waypoint, trying to thread the infinitesimal needle.



A rigorous treatment of the stochastic case, e.g.\ with a tube MPC and
quantile dynamics formulation \cite{2012-cannon-StochasticTubeMPC},
would require to make explicit quantile dynamics assumptions on the
system stochasticity and external perturbations. Such a treatment is
beyond the scope of this paper and we want to avoid formulating a
priori probabilistic assumptions about external perturbations.

We therefore propose a simple cutoff and backtracking approach: When
$\tau_1 \le \e$ we do not re-optimize the timing and let the system
instead continue to track the last spline reference $\xi^*$ until the
expected time of waypoint passage. Then, in the first MPC cycle after
the expected waypoint transitioning, we check whether the waypoint was
passed with sufficient accuracy to progress phase, or otherwise keep
the phase and old waypoint active, which automatically leads to a
looping spline and substantial increase of $\tau_1$ for re-targeting the
old waypoint. In the pseudo code, the cutoff is realized in line
\ref{codeCutoff}, and the waypoint check in line
\ref{codeWaypointFails}, or alternatively also in
\ref{codeRunningFails} which would backtrack if the subsequent running
constraint is missed and encodes the desired criteria for waypoint
collection.\footnote{If we had quantile assumptions on the system stochasticity and a
waypoint margin $\t$, we could choose $\e$ based on this, namely choose
$\e$ such that $P(|x(t+\e)-x^*(t+\e)|>\d) < \a$, i.e.\ the probability
of missing the reference with margin at time $t+\e$ is lower than a
given $\a$. By choosing $\a$ we choose how often (theoretically) the
system will fail and have to loop back. However, rather than making
explicit assumptions about system (and external) stochasticity, we
heuristically fix $\e$ to a time horizon, typically of 0.1 seconds.}





\subsection{Phase Backtracking \& Overall \secmpc{} Cycle}

We add phase backtracking to our \secmpc{} cycle (line
\ref{codeRunningFails}), which re-initiates the previous phase and
waypoint if the current running constraints are missed. Reassigning
$\t_\k\gets \tau_\text{init}$ means to reinitialize the delta timings
away from zero, so that subsequent timing optimization can better
converge.
Further, line \ref{codeFilter} allows to include a filter on the
actuated dofs to stabilize the system. In particular, if the inverse
dynamics used to execute $\xi^H$ has a systematic error (as is the
case for our used robotic system, a Franka Panda), an MPC that fully
adopts the true current state as the start state for a new reference
can be too compliant to the tracking error. Instead of
fully adopting the current state as reference start state, we choose a
reference start state in the interpolation between the old reference
point $\xi^*(\rt)$ and the true current state $x$, with at max
distance $r$ to the current state, where $r$ equals the typical joint
space tracking error of the system.

\section{Experiments}

\subsection{Reactive Pushing under Interventions}

We start with discussing a real-world pushing scenario that best
motivates our approach and highlights the properties of the \secmpc{}
system. Below we discuss two more real-world scenarios, pick-and-place
and quadrotor through gates, as well as studies on simplistic settings
to analyze \secmpc{} properties. Please see the accompanying video as well as raw video footage and
the fully open code for the following
experiments.\footnote{{\scriptsize\url{https://www.user.tu-berlin.de/mtoussai/22-SecMPC/}}}

We use Optitrack to sense the pose of objects and execute \secmpc{} on
a standard i8 core RT Linux machine without parallelization or use of
a GPU with a cycle time of 30msec. We choose a short horizon of
$H=1$sec and a time resolution of 100msec within the short horizon
sub-problem, and a cutoff of $\e=100$msec. Each sub-problem is
addressed using a 2nd-order Gauss-Newton method. The actually required
number of Newton steps can be high (above 50, in each of the three
sub-problems) at initialization or with significant perturbations, but
very low (below 5) when perturbations are low and the warmstarts are
close to the solution. In the latter case, the actual compute time for
a \secmpc{} cycle is less than 10msec; in the first case, the actual
compute time can exceed the 30msec and thereby slow down the MPC
cycle. Due to its general non-linearity, waypoint optimization is in
principle prone to converging to an infeasible point or aborting
(after 300 Newton steps). However, in our experimental setups we did
not experience local optima and this never occurred.

\begin{figure}\centering
  \showh[.35]{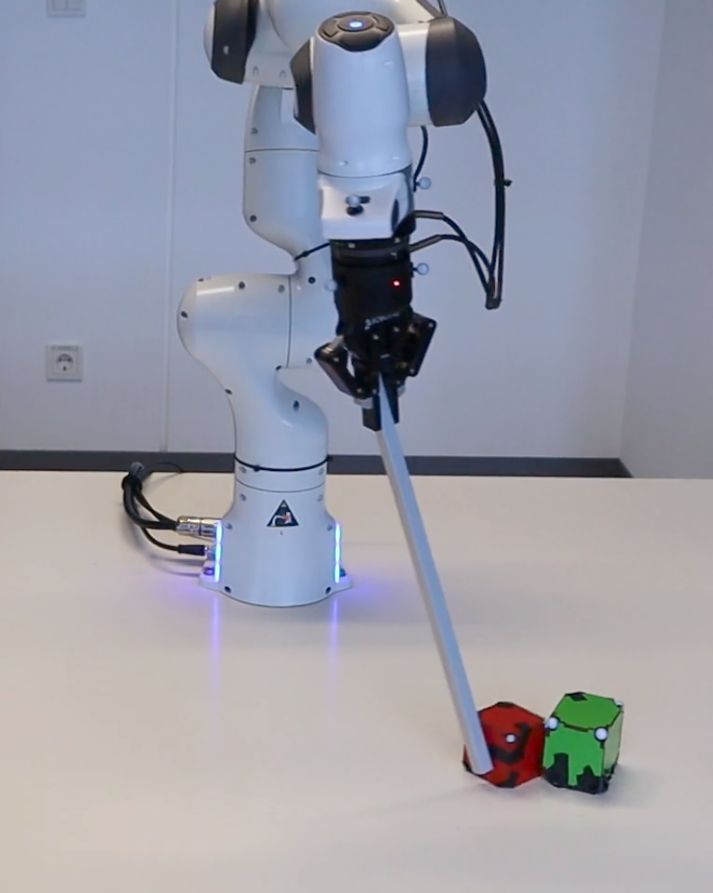}~
  \showh[.35]{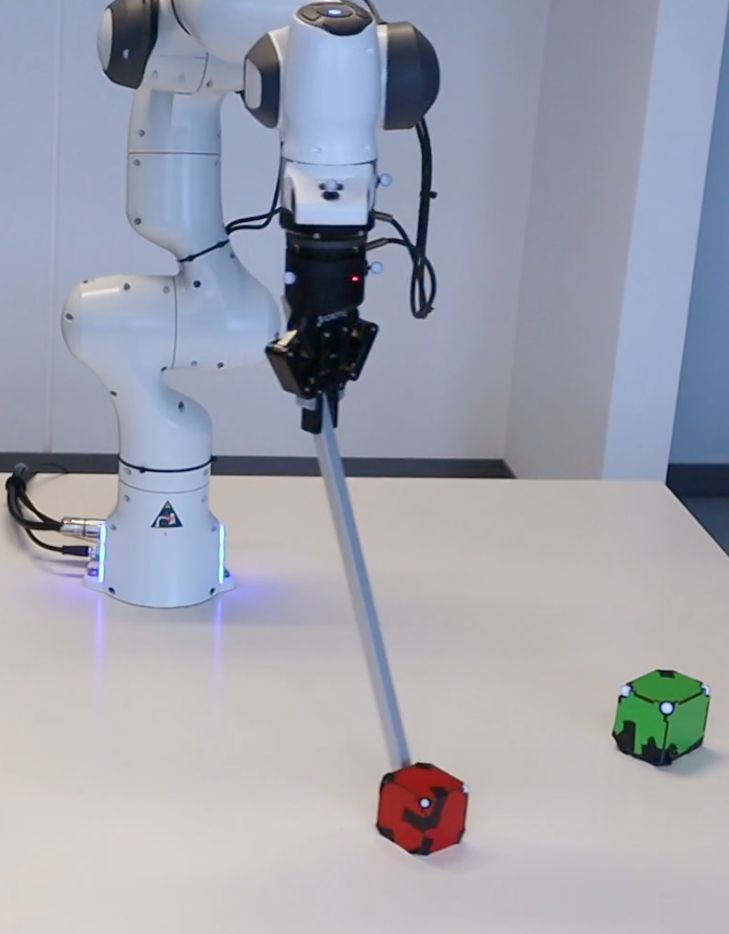}~
  \showh[.25]{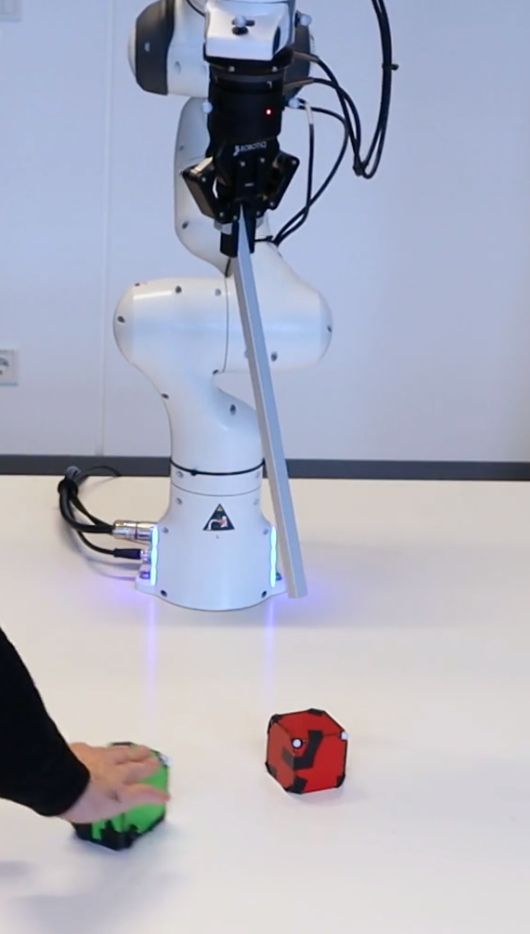}
  \caption{\label{figPushing} Pushing scenario, where the task is to push a red box into contact with a green (left). Fluent pushes frequently let the red box rotate out of the push (middle), and the experimenter interferes by replacing the target box during execution (right), both of which leads to robust re-initiation of the manipulation sequence.}
\end{figure}

Fig.~\ref{figPushing} shows a pushing scenario where a Panda robot
holds a stick 50cm long with which it aims to push the red block to establish contact with the green block. The potential perturbations in this scenarios are abundant:
\begin{enumerate}
  \item The red and green blocks can be replaced at any point during the execution by an interfering experimenter.
  \item The experimenter can manually hold back the robot arm, perturb it from its path, or manually bring a collision obstacle (additional stick) into the scene to further interfere with the execution.
  \item The real-world behavior of the red block under pushing contact
    is rather unpredictable.\footnote{The grip of the stick in the
      robot hand is not perfectly stable and imprecise; the red block
      pose estimation underlies stochasticity from Optitrack; the
      precise push contact is unknown; the precise ground-box
      interaction via an inhomogeneous and not perfectly even box
      lower side is uncertain.}
\end{enumerate}
Therefore, rather than hoping that we could have a (probabilistically) accurate model of
the dynamics, we aim for a system that is robust to such
interferences and failures of intermediate phases, in particular by
re-initiating the manipulation sequence as needed.

We model the pushing behavior as a sequence of four constraints, $\hat\phi_{1:4}$, where the first three are simple geometric constraints on approach waypoints, and the pushing itself occurs from the 3rd  waypoint. As the final target constraint is highly coupled to all previous constraints, we start explaining the constraints backward:
\begin{itemize}
\item $\hat\phi_4$ constrains the distance between red and green box to zero; this implies the waypoint solver to maintain an estimate of the final place pose of the red box.
\item $\hat\phi_3$ constrains the stick tip to be in contact with the red box at a position that is \emph{opposite} to the final place pose relative to the red box center -- this describes a central push on red toward the final pose.
\item $\hat\phi_2$ is almost identical to $\hat\phi_3$, but 3cm away from contact (modeling a pre-push pose waypoint); and $\hat\phi_1$ is similar to $\hat\phi_2$, but 10cm higher.
\end{itemize}
The pre-push pose constraints $\hat\phi_{1,2}$ imply approach
waypoints that can efficiently and fluently be transitioned by
\secmpc. Such approach constraints could have been designed in other
ways, e.g.\ imposing constraints on approach velocities or directions,
but we found this easiest and exploited \secmpc's fluency in
transitioning waypoints.

Throughout all phases we have the same running
constraints $\bar\phi_{1:4}$, which constrain the stick tip to be on
the opposite side to the final place pose relative to the red box
center, as well as constrain the stick tip, red box center, and final
placement to be aligned on a single line. Whenever these running
constraints are missed, \secmpc{} will initiate phase backtracking.

\secmpc{}
achieves a highly robust behavior capable to cope with the mentioned
interferences and perturbations. In particular, without interference
by the experimenter, the system shows fluent push motions that are
aborted and re-initiated whenever the red object rotates out of the
push, robustly leading to fulfilling the task. When the experimenter
interferes by replacing the green box, this changes the final target
constraint $\hat\phi_4$ and thereby all waypoint and running
constraints, typically also leading to a phase backtracking and
re-initiation of the maneuver to ensure alignment to an updated red
box target pose.

\subsection{Reactive Pick-and-Place}

\begin{figure}\centering
  \showh[.34]{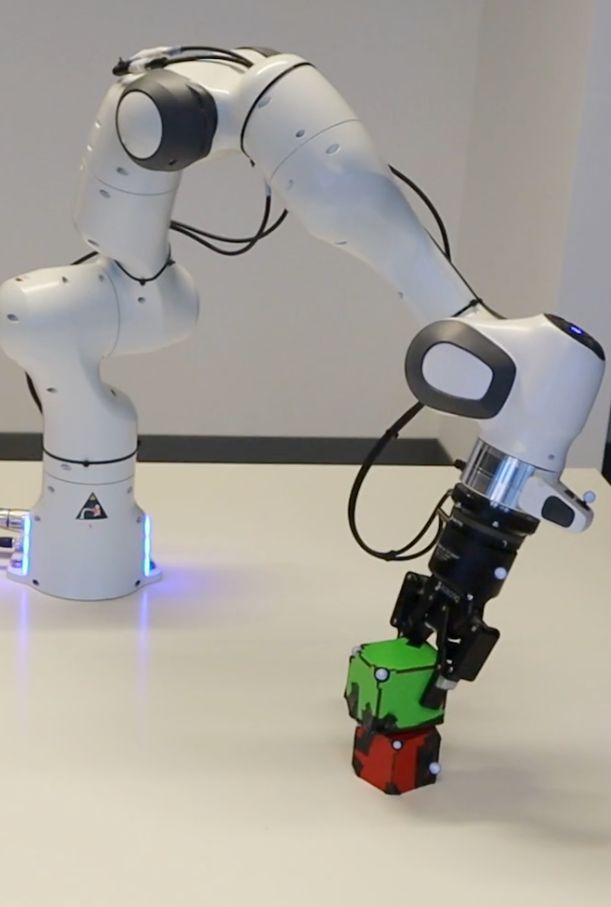}~
  \showh[.32]{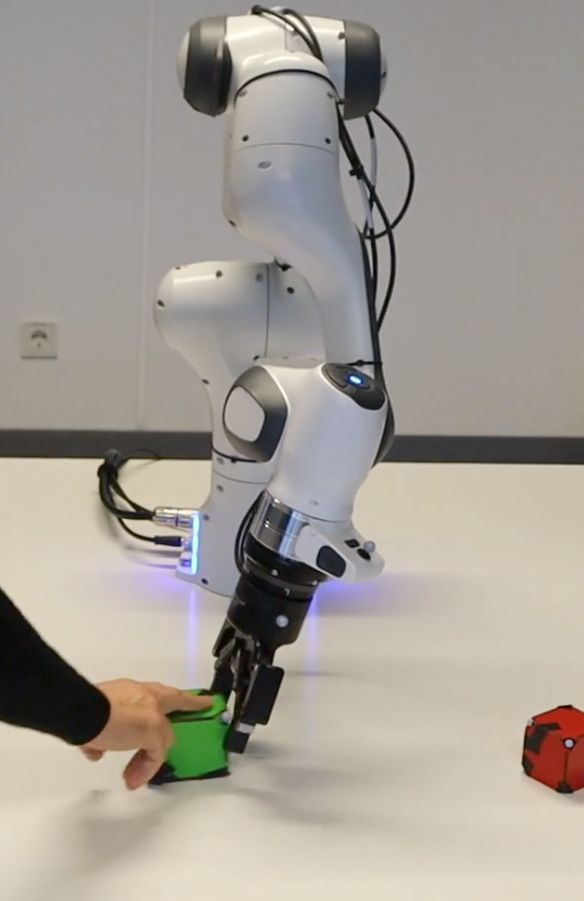}~
  \showh[.28]{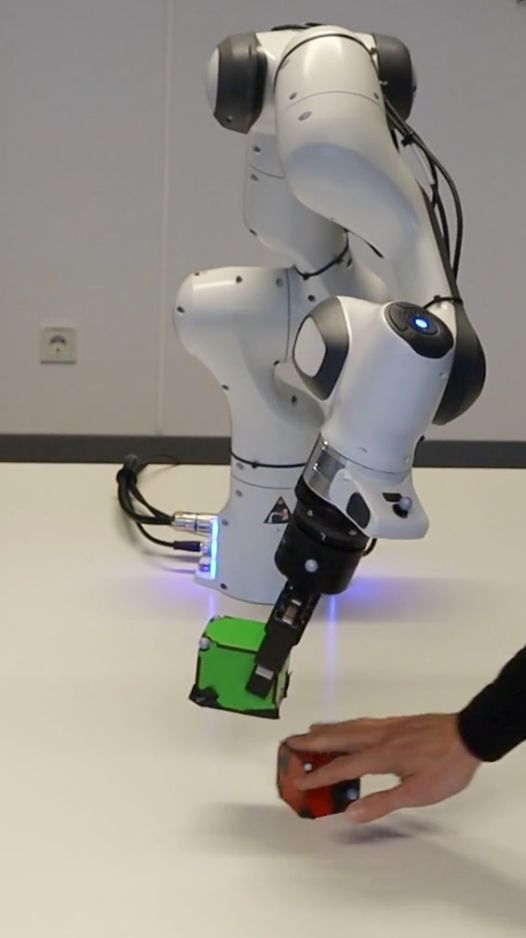}
  \caption{\label{figPnp} Pick-and-place scenario, where the task is to stack a red box onto a green (left). The experimenter interferes by stealing, replacing and rotating the box during pre-grasp (middle), and replacing the target during pre-place (right).}
\end{figure}

In a second scenario we consider pick-and-place under interferences,
see Fig.~\ref{figPnp}. Here, the constraints and geometry were
modeled exactly as in previous work on TAMP \cite{2018-toussaint-DifferentiablePhysicsStable}, showing that
TAMP planning models can directly be transferred to a reactive
execution system using \secmpc. Again, the system exhibits robustness
to interferences by the experimenter when stealing and replacing the
box during pre-grasp, replacing the target during placement, or
directly pushing the arm or intervening with a manual obstacle
stick. For brevity, we refer to the accompanying video to showcase
this scenario.

\subsection{Quadrotor through Gates}

\begin{figure}\centering
  \showh[.44]{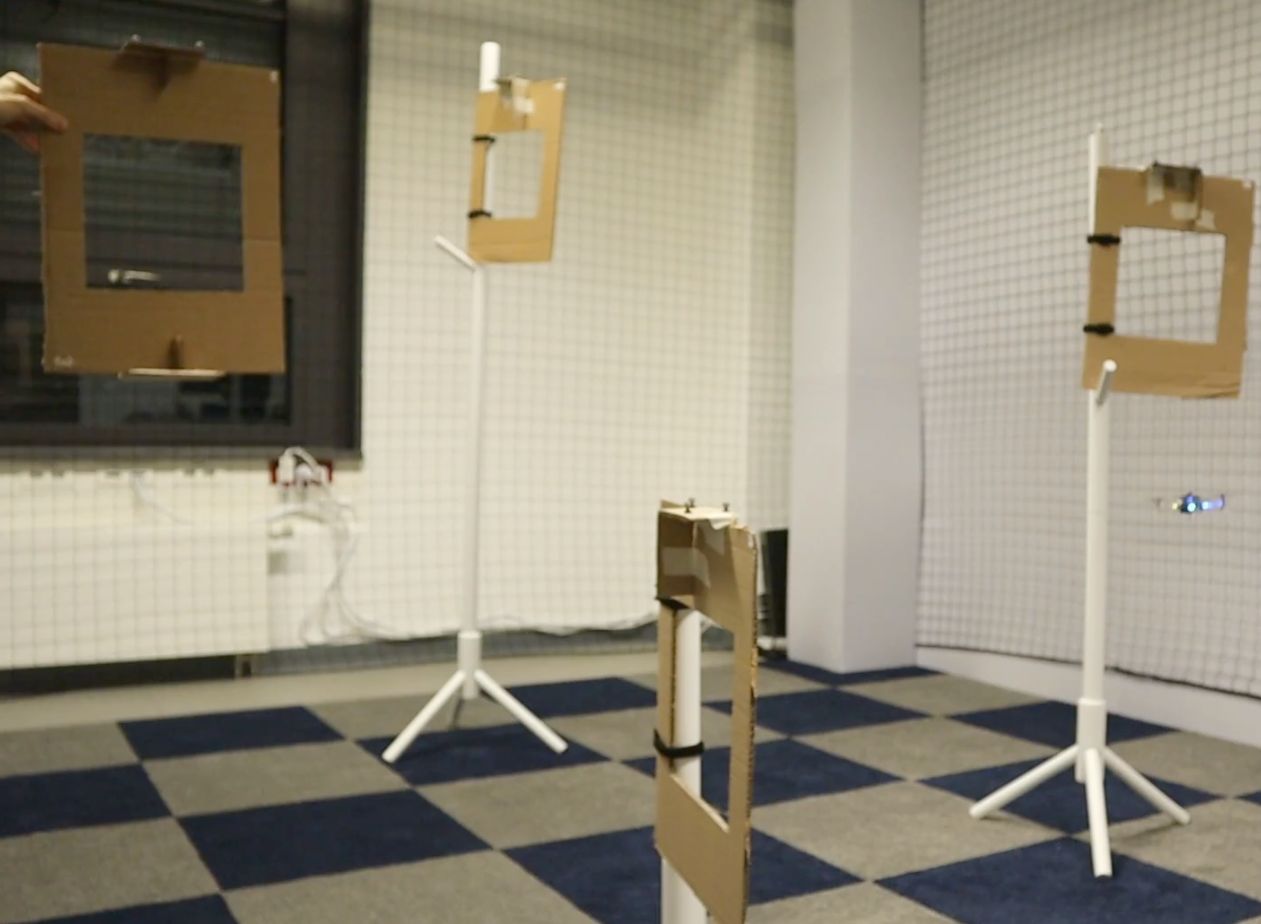}~
  \showh[.43]{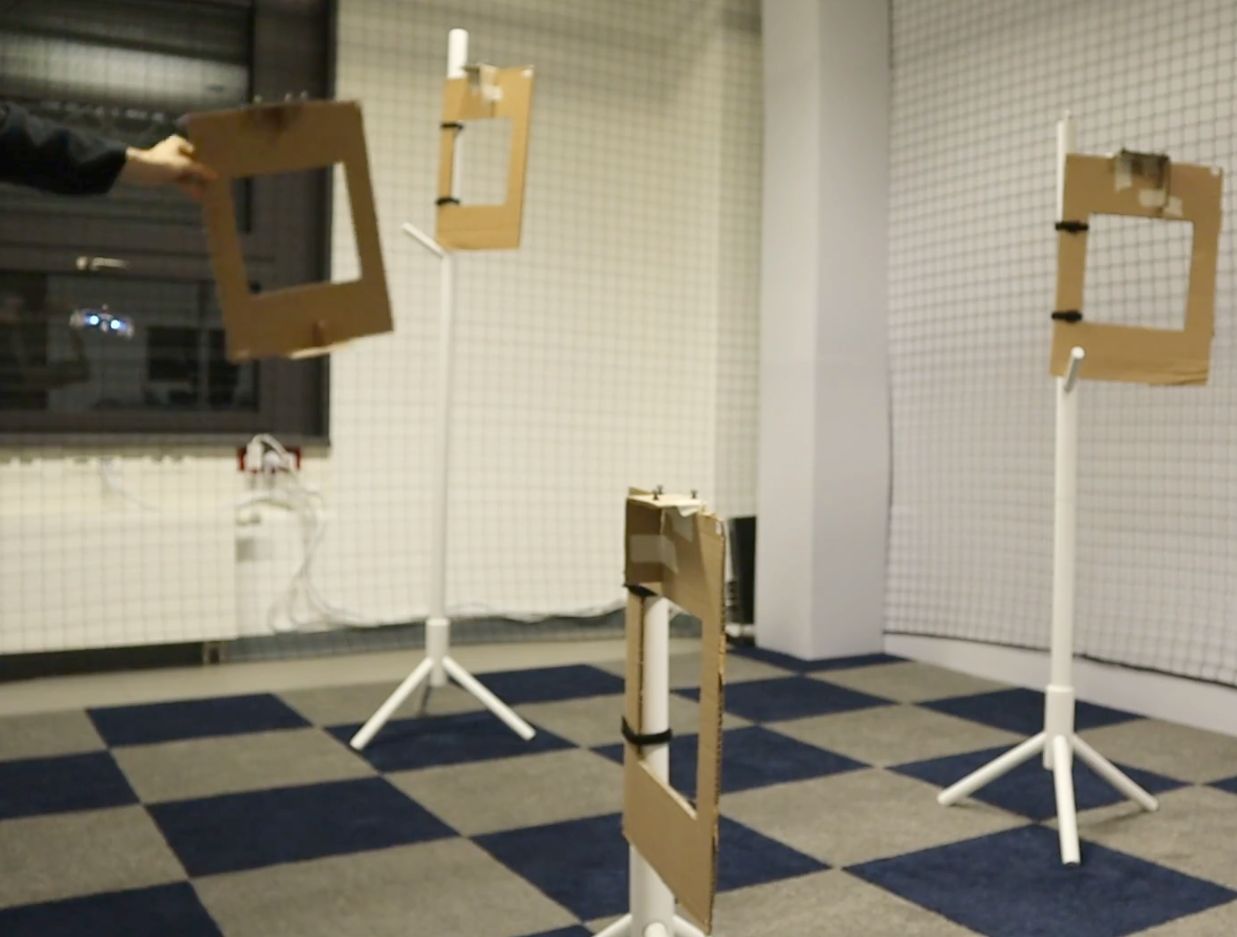}
  \caption{\label{figQuat} Quadrotor scenario, where the task is to (indefinitely) transition through the gates (left). The experimenter interferes by displacing the left gate during the approach (right).}
\end{figure}

To highlight that our system is not specific to robotic
manipulation, we demonstrate \secmpc{} also on a quadrotor scenario,
see Fig.~\ref{figQuat}. Again, for brevity we refer to the
accompanying video to showcase this scenario.

\subsection{Analysis on Analytic Settings}

\begin{figure}\centering
  \twocol{.43}{.55}{ \showh[1]{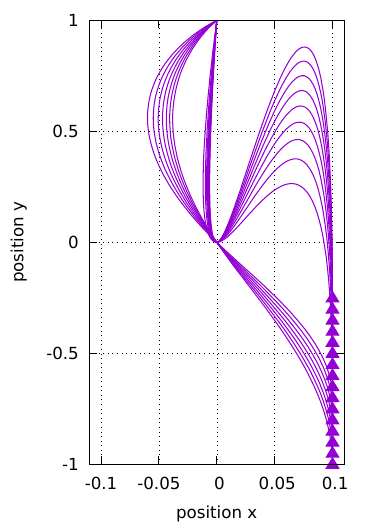}\anchor{-100,-5}{(a)} }{
    \showh[1]{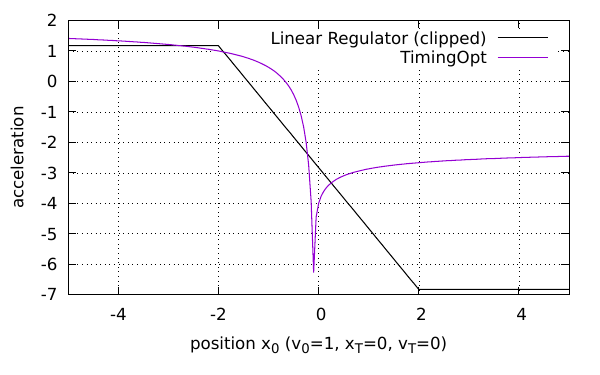}\anchor{-130,5}{(b)}\\ \showh[1]{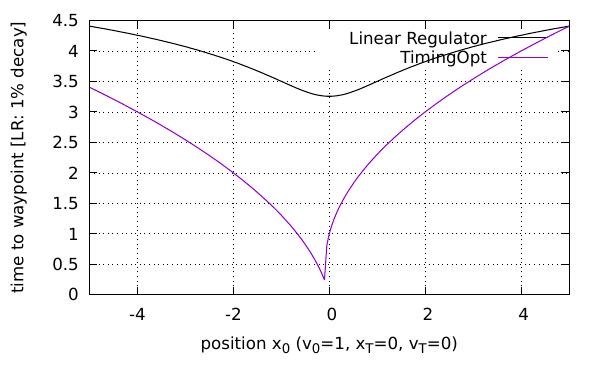}\anchor{-130,5}{(c)}
  }
  \caption{\label{figTO} (a) Timing-optimized trajectories in 2D through two waypoints $(0,0)$ and $(0,1)$, starting with upward velocity $(0,1)$, slight $x$-offset, and different $y$-offset (triangles are starting points).
    (b) Acceleration of TimingOpt vs.\ a critically damped clipped Linear Regulator, when starting in 1D at $(x,\dot x=1)$ to waypoint $(x=0,\dot x=0)$.
    (c) Exact time-to-go for TimingOpt vs.\ time-to-1\%decay for the Linear Regulator}
\end{figure}  

We add results on simplistic low-dimensional settings to gain more
insights in the timing-optimal behavior of \secmpc.
Fig.~\ref{figTO}(a) displays timing-optimal trajectories (solving
(\ref{eqTimingOrg})) in 2D configuration space for two aligned
waypoints. When the system is close to the first waypoint with straight
velocity but lateral offset, we see the discontinuous switch in
deciding to pass the waypoint and loop back (Sec.~\ref{secStoch}).

Figs.~\ref{figTO}(b,c) compare the convergence to a single waypoint in a 1D
configuration space to a critically damped regulator.
A naive linear regulator would have undesirable high feedback for
large initial error; a clipped regulator (where the error or feedback
is upper bounded, as typically used in practice) would have acceptable
behavior far from the waypoint, but still ``only'' exponential
convergence near the waypoint. A timing-optimal MPC approach to the
waypoint generates moderate feedback far and near the waypoint while
at the same time providing an exactly timed and definite convergence
to the waypoint if no perturbations.

We also compared the average total time to transition through 5
perturbed waypoints when using \secmpc{} through the full sequence
vs.\ sequencing five independent 1-phase approaches. Each 1-phase approach is realized
with \secmpc{} that only knows about the next waypoint (which is already an
improvement over using a linear regulator to individually approach
each waypoint, as shown above). As expected, when waypoints are
somewhat aligned, the full \secmpc{} can fluently transition them and requires
significantly less total time for transition. For fully random
waypoints $\sim [-1,1]^3$, we still found a reduction in total time
from $14.1\pm0.6$sec to $11.2\pm0.6$sec, averaged over 20 random
trials.

\section{Conclusion}

In this paper we derived a reactive control strategy to execute a
given linear sequence of constraints. The scope of this paper is more
narrow than previous work on full online replanning in a TAMP setting
\cite{2019-paxton-RepresentingRobotTask,2021-li-ReactiveTaskMotion,2020-stouraitis-OnlineHybridMotion}. However,
the focus on a given sequence of constraints allows us to derive a
control strategy that encompasses the interdependencies between future
waypoints and their timing and includes timing optimization and phase
backtracking as an integral part of reactiveness. A core limitation of
the approach arises from the approximate decomposition of the full
problem, in particular from representing the long term path only
coarsely with waypoints. Running constraints between future waypoints, such
as collision avoidance, are therefore not evaluated for waypoint
estimation, but only accounted for in the short receeding horizon
path.


On a more conceptual level, it is a standard paradigm to think of long
term manipulation behavior as composed of individual primitive
controllers, or motor skills, that are sequenced e.g.\ by RLDS
\cite{2019-paxton-RepresentingRobotTask}, other state machines
\cite{2000-egerstedt-BehaviorBasedRobotics}, or tree structures
\cite{2017-majumdar-FunnelLibrariesRealtime}, and where each primitive
has its own control law, e.g.\ acting as a funnel, or option (in the
context of hierarchical RL). The presented system provides an
alternative view, where not individual control laws are the basic
building blocks, but constraints. Compositionality (and decision
abstraction) is achieved on the level of constraints, and instead of
composing primitive control laws, here we derive control directly from
the composition of constraints. The constraint-based view on control
is not novel \cite{2014-aertbelien-ETaSLETCConstraintbased}, but our
system provides an explicit timing-optimal MPC realization for a
sequential composition of constraints.





\bibliographystyle{IEEEtran}
\bibliography{group,22-SecMPC}
\end{document}